 \documentclass[tablecaption=bottom,pmlr]{jmlr}

\usepackage{booktabs}
\usepackage[utf8]{inputenc} %
\usepackage[T1]{fontenc}    %
\usepackage{hyperref}       %
\usepackage{url}            %
\usepackage{amsfonts}       %
\usepackage{nicefrac}       %
\usepackage{microtype}      %
\usepackage{xcolor}
\usepackage{tikz}
\usepackage{aas_macros}
\usepackage[normalem]{ulem}

\usepackage{amsmath}
\usepackage{amssymb}
\usepackage{physics}

\usepackage{xcolor}

\usepackage[load-configurations=version-1]{siunitx} %
\clearpage{}%

\newcommand\bv{\mathbf{v}}

\makeatletter
\def\BState{\State\hskip-\ALG@thistlm}
\makeatother

\setcitestyle{square,numbers}

\clearpage{}%
\theorembodyfont{\upshape}
\theoremheaderfont{\scshape}
\theorempostheader{:}
\theoremsep{\newline}

\jmlrvolume{1}
\jmlryear{2021}
\jmlrsubmitted{submission date}
\jmlrpublished{publication date}
\jmlrworkshop{NeurIPS 2020 Preregistration Workshop} %

\title[Unsupervised Resource Allocation with GNNs]{Unsupervised Resource Allocation\\with Graph Neural Networks}

  \author{
  \Name{Miles Cranmer\nametag{\thanks{Corresponding Author}}} \and \Name{Peter Melchior}\\
   \addr 
    Princeton University\\
    Princeton, NJ 08544, USA\\
    \Email{\{mcranmer,peter.melchior\}@princeton.edu}\\
  \AND
  \Name{Brian Nord} \\
  \addr Fermilab \\ Batavia, IL 60510, USA\\
  \Email{nord@fnal.gov}
}

\newcommand\blfootnote[1]{%
  \begingroup
  \renewcommand\thefootnote{}\footnote{#1}%
  \addtocounter{footnote}{-1}%
  \endgroup
}

\begin{document}

\maketitle

\begin{abstract}
We present an approach for maximizing a global utility function by learning how to allocate resources in an unsupervised way.
We expect interactions between allocation targets to be important and therefore propose to learn the reward structure for near-optimal allocation policies with a GNN.
By relaxing the resource constraint, we can employ gradient-based optimization in contrast to more standard evolutionary algorithms.
Our algorithm is motivated by a problem
in modern astronomy, where one needs to select---based on limited initial information---among $10^9$ galaxies those whose detailed measurement will lead to optimal inference of the composition of the universe.
Our technique presents a way of flexibly learning an allocation strategy
by only requiring forward simulators for the physics of interest and the measurement process.
We anticipate that our technique will also find applications in a range of allocation problems from social science studies to customer satisfaction surveys and exploration strategies of autonomous agents.
\end{abstract}


\section{Resource Allocation for Observational Studies}
We consider a problem that frequently arises in observational sciences as well as the training of autonomous agents. \blfootnote{Project repository at \url{https://github.com/MilesCranmer/gnn_resource_allocation}}
Due to resource limitations, not every desirable action is admissible, leading to the problem of finding a strategy that makes best use of the available resources. 
We are particularly interested in a scenario where many individual actions need to be performed before their combined utility can be determined.
In such a scenario it should be beneficial to learn not just which actions have high reward but also how interactions between possible actions influence the outcome.

\begin{figure}[t]
    \centering
    \includegraphics[width=0.98\textwidth]{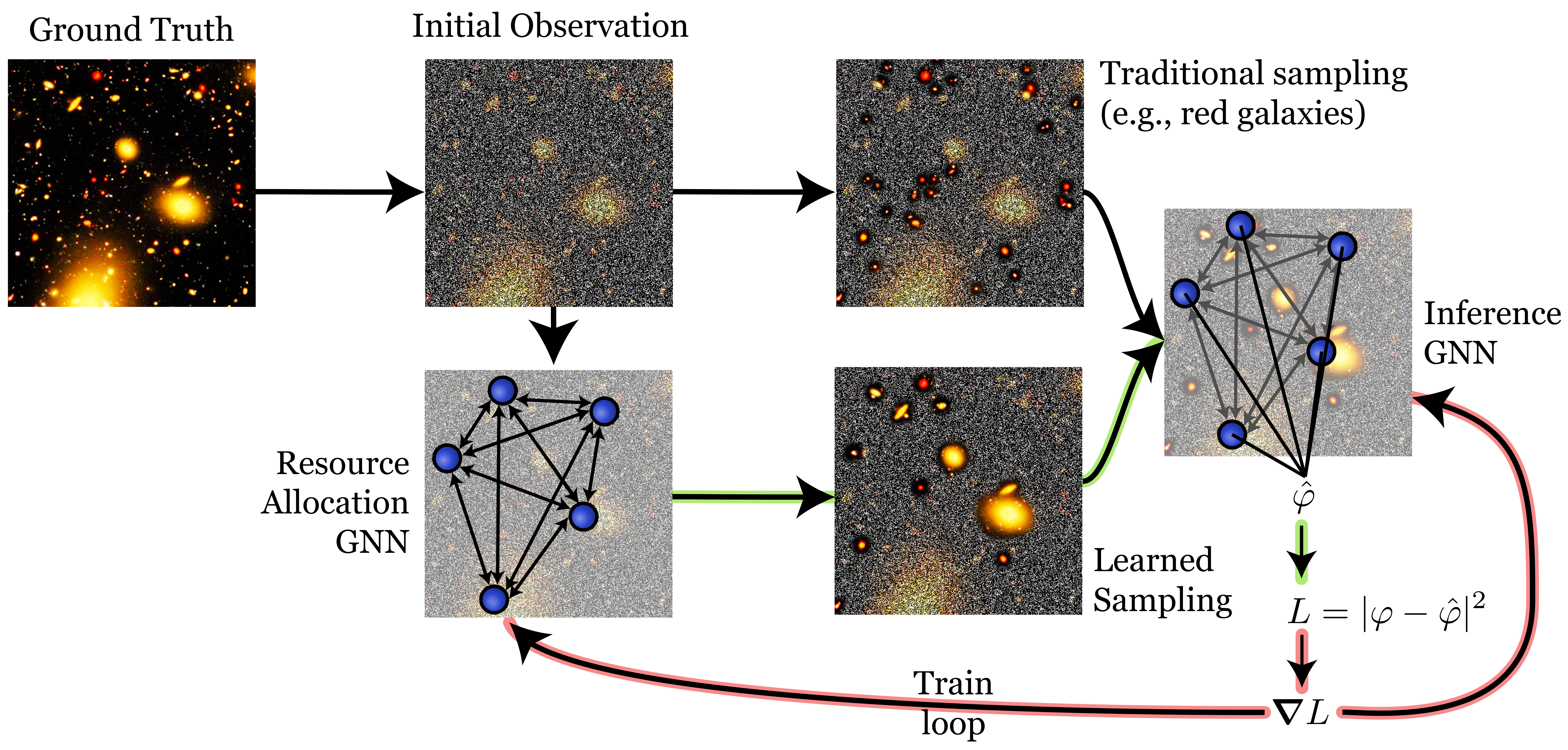}
    \caption{Schema of the proposed technique applied to an astronomical dataset.
    Given initial (noisy) information, $\textit{GNN}_1$ proposes how the distribute observing resources over all galaxies, after which $\textit{GNN}_2$ predicts a more accurate posterior estimate of a global variable $\varphi$ of scientific interest. The accuracy of this prediction is used to update both GNNs to better allocate resources. A traditional allocation method is shown only for comparison (see \autoref{sec:hparams}).
    }
    \label{fig:demo}
\end{figure}

As a concrete example, modern astronomical observations fall into two groups. 
So-called surveys cover wide areas of the sky without any specific selection of celestial objects.
On the other hand, targeted observations often draw from prior surveys to select specific objects for detailed characterization.
Telescope facilities for targeted studies are rare and expensive to operate.
Access to them is severely limited and competitively allocated based on the perceived merit of the proposed scientific study.
After decades of astronomical sky surveys, \emph{some} information is available for hundreds of millions of celestial objects (such as their position on the sky and their brightness), but detailed information can only be acquired for a much smaller subset.
Based on what is already known of these objects, astronomers need to decide which of them merit further study (see \autoref{fig:demo} for a schema of the problem).
The situation resembles that of an organization whose management wants to incentivize beneficial user behavior through an intervention aimed at specific users; or of an autonomous agent that has a ``foggy'' view of its surroundings and only enough sensory bandwidth to explore a few possible options before having to move on.

We are interested in finding a policy for taking actions in an unknown environment $M\in\mathcal{M}$, described as a Markov Decision Process (MDP), $M=(\mathcal{S}, \mathcal{A}, \mathcal{P}, \mathcal{R}, H)$, where $\mathcal{S}$ is the set of possible states, $\mathcal{A}$ the set of actions, and $\mathcal{P}$ the set of transition probabilities from  state $s$ to state $t$ by means of action $a$, $\mathbb{P}(t\mid s,a)$, resulting in the reward $r(s, a)\sim \mathcal{R}$. The horizon $H$ describes how many actions can be taken and reflects the resource limitations.
In contrast to problems in scheduling and control, our problem is not sequential and can thus be described as finding the policy
\begin{equation}
\label{eq:policy}
    \pi^* = \mathrm{argmax}_\pi\ U(s_H^\pi)
\end{equation}
for some domain-specific utility function $U$.
Note that $U$ depends only on the final state $s_H^\pi$ after all $H$ actions have been taken according to policy $\pi$, but not on the ordering of the actions.

For observational sciences, the state set is given by all possible study subjects $\bv_i\in\mathbb{R}^{n_v}$, $i=1\dots N$, more precisely what we know about them. 
Possible actions are restricted to allocating an available resource to observe a subject $i$, which amounts to advancing our knowledge from a prior $p(\bv_i)$ to a posterior $p(\bv_i\mid a_i = \mathrm{observe}\ i)$.
The set of observed subjects $\mathcal{O}^\pi = \lbrace i: a_i = \mathrm{observe}\ i\rbrace$ has cardinality $H$.
The final state is thus given by 
$s_H^\pi = \lbrace p(\bv_i)\rbrace_{i\notin\mathcal{O}^\pi} \cup \lbrace p(\bv_i\mid i\ \mathrm{observed})\rbrace_{i\in\mathcal{O}^\pi}.$    
The utility is then usually the inverse variance of the posterior of parameter $\varphi$ for some hypothesized model of the process under study, based on the collection of observations,
\begin{equation}
\label{eq:utility}
 U(s_H^\pi) = \mathrm{var}^{-1}\left[p(\varphi\mid s_H^\pi)\right].
\end{equation}

The non-sequential nature of this problem allows a substantial simplification.
Because any action amounts to observing one $i\in\lbrace 1,\dots,N\rbrace$, the reward $r$ can be expressed on per-subject basis:
$r(s,a = \mathrm{observe}\ i) = r(i\mid\mathcal{O})$,
i.e. the reward of adding $i$ to $\mathcal{O}$.
We seek to determine the reward for every possible subject such that the final set $\mathcal{O}$ maximizes $U$.
We propose to parameterize $r(i\mid\mathcal{O})$ as a Graph Neural Network (GNN) $r_\theta(\lbrace \bv_i\rbrace)$ to exploit possible relations between subjects in forming the optimal set $\mathcal{O}$. 
GNNs, as defined in \citep{Battaglia2018-cy}, are neural networks
that accept as input or output an arbitrary graph (defined as a set of nodes and edges) or a scalar variable.
A GNN can be thought of as the generalized version of an Interaction Network (INs) \citep{battaglia2016interaction}, using inductive biases motivated from physics.
Early work on GNNs include \citep{scarselli2009graph,bronstein2017geometric,gilmer2017neural}.
The optimal policy can be determined by rank-ordering $r_\theta(\lbrace \bv_i\rbrace)$ and selecting the highest-ranked $H$ subjects.
The advantage of this approach lies in avoiding black-box optimization (by e.g. evolutionary algorithms) usually employed for finding solutions to \autoref{eq:policy} \citep{Storn1997-he, Naghib2019-ko}.
Instead, we can utilize novel approaches of black-box back-propagation \citep{Vlastelica2019-np, Tabibian2020-cn} or perturbed optimization \citep{Berthet2020-co} for combinatorical problems.

We can further relax the strict set assignment $i\in\mathcal{O}$ and interpret the prediction of the GNN as the \emph{amount} of allocated resources for each subject, e.g. observing time, so that $\sum_i r_\theta(\lbrace \bv_i\rbrace) = H$. This redefinition allows us to put the sum constraint into the loss function.
If a non-zero minimum allocation is required for any improvement in $U$ (as is often the case in practice), the network predictions will naturally approximate the previous case of a small number of subjects being observed.
We continue to employ gradient-based optimization to learn the resource allocation with a GNN and replace the black-box relaxation hyper-parameter in \citep{Vlastelica2019-np, Berthet2020-co} with a variable feasibility penalty on $H$, increasing it until the constraint is satisfied within acceptable bounds.

\section{Context \& Related Work}

The main contributions of our proposal are two-fold. First, by gradually transitioning from an unconstrained problem to a constrained problem, it should be easier for the GNN to first learn which samples are relevant for maximizing the utility, and then to adjust to the resource limitations --- all within the same gradient-based architecture.
Second, whereas traditional approaches often rely on manually defined reward functions based on heuristics, we learn a reward function from scratch with the help of a second GNN and forward simulations of the parameter(s) of interest. This flexibility should allow for further improvements in the allocation strategy.
Furthermore, performing symbolic regression on the GNN \citep{cranmergn1,cranmergn2} can lead to generalizable formulations of interaction terms that reveal which nodes of the subject network need to be maintained under resource limitations. Providing such a form of model interpretability is very important to the astronomical community for their adoption of any proposed technique.

Resource allocation through Reinforcement Learning (RL) is a rich problem space that spans a wide range of activities and domains. These include control of and communication between autonomous agents (e.g., self-driving cars), assignment of tasks and resources to nodes in grid computing, communication across wireless networks, as well as selecting study subjects in the observational sciences \cite{RL1,RL2,RL3,8633948}.
Topically most similar to our work, \cite{Naghib2019-ko} optimized the observation schedule for a large astronomical survey with an evolutionary algorithm.

GNNs in particular have been used in prior work for learning policies.
The ``NerveNet'' model \citep{nervenet} uses message passing between different body parts to determine a generalizable policy for agent movement.
The ``policy-GNN'' \cite{policygnn} uses similar principles to our proposed architecture, using GNNs to select the next node in a graph, but considers meta-learning for traditional GNN classification algorithms rather than resource allocation, and cannot be reconfigured to our problem. It also assumes iterative sampling, whereas for our astronomical problem one needs to select the samples at once, based on limited information about the graph. 
Additional examples of GNNs used for resource allocation include \cite{gnnalloc,gnnalloc2}, which were applied to resources on a wireless network and focus on the edges of a GNN, whereas we use the edges merely as a way of factoring in interactions between potential subjects.

\section{Methodology \& Experimental Protocol}

We motivate our experiment in \autoref{sec:study} by a real-world problem from current astronomical research. In \autoref{sec:arch}, we give a detailed definition of our GNN variant, with a full list of hyperparameters, and the algorithm to train it. In \autoref{sec:hparams}, we define how we will perform the experiments, the metric for success, as well as the comparison to existing allocation policies.

\subsection{Case Study}
\label{sec:study}

Motivated by a challenging problem for large-scale astronomical surveys of the coming years, we focus on the following case study:
the selection of celestial objects, usually called ``targets'', to improve our knowledge of the make-up of the universe.
In particular, we will look at the measurement of $O(10^9)$ distant galaxies, for which very accurate positions on the sky $(x_1,x_2)$ are known, but their distances $d$ and masses $m$ are known only to within 10-30\%.
For precise cosmological studies, all four features, $\bv=(x_1,x_2,d,\log m)$, need to be known accurately, the distance to within $\approx 0.1\%$.
As astronomers are able to perform targeted observations for $O(10^7)$ galaxies \citep{Levi2019-ws}, we need to decide which galaxies should be further studied with a more accurate instrument.
The final utility from \autoref{eq:utility} will be given by the inverse variance on a single parameter $\varphi=\Omega_m$, the fraction of the total energy density in the universe in the form of matter (as opposed to e.g. radiation).%
\footnote{The interested reader can find more information about cosmology in \citep{Coles2001-ql}.}
Broadly speaking, increasing $\Omega_m$ increases the amount of spatial clustering of galaxies throughout the universe. Thus, knowing their precise 3D positions and masses allows to recover $\Omega_m$.

\newcommand\R{\mathbb{R}}
\newcommand\rtr[2]{\R^{#1}\rightarrow\R^{#2}}
\subsection{Architecture \& Algorithm}
\label{sec:arch}
We propose to combine two independent GNNs: $\textit{GNN}_1$ encodes the reward, or rather: the resource allocation for every target in a simulated universe \citep{Villaescusa-Navarro2020-wa}, as a set-to-set mapping.
$\textit{GNN}_2$ seeks to estimate the cosmological parameter of the simulated universe, thereby emulating the posterior inference step of \autoref{eq:utility} as a set-to-scalar mapping.
$\textit{GNN}_2$ is a fast proxy for a scientific analysis, which traditionally would take days or months to complete, and resembles the function of the value network in \cite{Silver2016-qe}.
We point out that the training of $\textit{GNN}_1$ is unsupervised as it is never shown a predefined optimal target set.
Our approach is visually depicted in \autoref{fig:demo}.

Our GNN models take the form of the full graph network block
as described in \cite{Battaglia2018-cy}, containing a global
block, node block, and edge block. We use three message passing steps, with
three separate ``blocks.''
Using notation from \cite{Battaglia2018-cy}, we denote multi-layer perceptrons (MLPs) by $\phi:\mathbb{R}^{(\cdot)}\rightarrow\mathbb{R}^{(\cdot)}$. Define
$\textit{GNN}_1$ as $\{\phi_1^{v,\text{enc}},\phi_1^{e,\text{enc}}, (\phi_1^{e,i},\phi_1^{v,i},\phi_1^{u,i})_{i=1\text{:}3}, \phi_1^{v,\text{dec}}\}$,
which maps from set to set. Here, 
$\phi_1^{v,\text{enc}}:\rtr{2}{n_v}$, which encodes mass and distance,
$\phi_1^{e,\text{enc}}:\rtr{2}{n_e}$, which encodes relative angular distance
between galaxies,
$\phi_1^{e,i}:\rtr{2n_v+n_e+n_u}{n_e}$,
$\phi_1^{v,i}:\rtr{n_v+n_e+n_u}{n_v}$,
$\phi_1^{u,i}:\rtr{n_v+n_e+n_u}{n_u}$, for each $i$,
which update the graph's edges, nodes, and global, respectively, and
$\phi_1^{v,\text{dec}}:\rtr{n_v}{1}$, which decodes
the final node state to assign resources.
The global vector is zero-initialized.
We define
$\textit{GNN}_2$ as $\{\phi_2^{v,\text{enc}},\phi_2^{e,\text{enc}}, (\phi_2^{e,i},\phi_2^{v,i},\phi_2^{u,i})_{i=1\text{:}3}, \phi_2^{u,\text{dec}}\}$,
which maps from set to scalar. The only architectural difference between these is
$\phi_2^{u, \text{dec}}:\rtr{n_u}{1}$, which predicts $\Omega_m$. All
aggregation operations are sum-pools. The sizes of the node
encodings are $n_v$, edges $n_e$, and global $n_u$, which are
hyperparameters. We define the adjacency matrix
based on k-nearest neighbors (with relative angular position
used as distance), with $k$ set as a hyperparameter.
These GNNs are to be trained using the procedure defined in \autoref{alg:gnn}.

\begin{algorithm2e}[t]
\SetAlgoLined
\SetAlgoNoEnd
\DontPrintSemicolon
\SetKwInOut{Input}{Input}\SetKwInOut{Output}{Output}
\Input{$H$ (total resource allocation), $\eta$ (resource allocation tolerance), $\lambda$ (learning rate), $\alpha$ (sparsity penalty)}
\Output{$\textit{GNN}_1$, the optimized survey scheduler; $\textit{GNN}_2$, the optimized inference statistic}
\BlankLine
$\theta_1 \gets \text{Kaiming initialization}$\;
$\theta_2 \gets  \text{Kaiming initialization}$\;
$(\tau,\delta\tau) \gets (0, 0.1 H^{-2})$\;
\While{$L$ not converged}{
    $\varphi \sim p(\varphi)$ \tcp*{parameter of simulated environment}
    $\{\mathbf{v}_i\} \sim \text{simulator}(\varphi)$ \tcp*{simulated data, arbitrary cardinality}
    \ForAll{$i=1\ldots|\{\mathbf{v}_i\}|$}{ 
        $\epsilon \sim \mathcal{N}(0, \Sigma_\mathrm{prior})$\;
        $\mathbf{v}'_i\gets \mathbf{v}_i+\epsilon$ \tcp*{simulated prior state}
    }
    $\{r_i\} \gets \textit{GNN}_1(\{\mathbf{v}'_i\}; \theta_1)$ \tcp*{resource allocation}
    \ForAll{$i=1\ldots|\{\mathbf{v}_i\}|$}{
        $\epsilon \sim \mathcal{N}\left(0, \Sigma_\mathrm{posterior}(r_i)\right)$\;
        $\mathbf{v}''_i\gets \mathbf{v}_i+\epsilon$ \tcp*{simulated posterior state}
    }
    $\hat{\varphi} \gets \textit{GNN}_2(\{\mathbf{v}''_i\}; \theta_2)$ \tcp*{parameter estimation}
    $L \gets (\hat{\varphi} - \varphi)^2 + \tau \left(\sum_i r_i - H\right)^2 + \alpha \sum_i \abs{r_i}$  \tcp*{combined losses}
    $(\theta_1, \theta_2)\gets  (\theta_1, \theta_2) - \lambda \cdot \nabla_\theta L$\;
    \If(){$|\sum_i r_i - H| > \eta$}{
        $\tau \gets \tau + \delta\tau$   \tcp*{allocation feasibility}
    }
}
\caption{Pseudocode for learning a strategy of allocating a finite amount of resources $H$ over a large set of study subjects. The goal is to advance from a prior (noisy) to a posterior (accurate) state so that the parameter of interest $\varphi$ can be optimally inferred.}
\label{alg:gnn}
\end{algorithm2e}

\subsection{Experiments}
\label{sec:hparams}

Mimicking the general conditions of a follow-up study with a spectrograph instrument at one of the world's largest telescopes, we standardize all four features to the range of $[0,1]$, and set the prior variances as $\Sigma_\mathrm{prior}(\bv_i) = \mathrm{Diag}(0,0,0.1,0.25)$.
The posterior variance dependents on the amount of resources $r_i$ spent on target $i$ as well as its mass and distance because more massive nearby galaxies are brighter and thus easier to observe in general:
\begin{equation}
\label{eq:posterior}
    \Sigma_\mathrm{posterior} = \begin{cases}
    \Sigma_\mathrm{prior} & \text{if}\ \ r_i < r_\mathrm{min}(d, \log m)\\
    \mathrm{Diag}\left(0,0,0.001,0.1\right) & \text{else.}\\
    \end{cases}
\end{equation}
The minimum resource requirement reflects that observations below a certain threshold are practically useless or impossible to perform; it is typically determined by the scientific study itself. We will adopt $\bar{r}_\mathrm{min}/H=10^{-7}$.
Its presence is also meant to ensure that the allocations are sparse. 
As long as $r_\mathrm{min}$ is small, the presence of solution polytopes should be tolerable during training.
If not, or if the allocations remain insufficiently sparse, we will add a $\ell_1$ penalty term to the loss function, whose strength is treated as a hyperparameter.

For realistic surveys, the outcome depends only weakly on the exact value of $H$, with practical influences, such as weather, likely to play a more important role. Consequently, we have some flexibility $\eta$ in the total allocation and will set $\eta/H=10^{-3}$.
As this target might be easier to meet, we will experiment with a fixed $\tau$ instead of a variable one, and treat its value again as a hyperparameter.

We will do a coarse hyperparameter search over internal
architectures of the MLPs for both GNNs: specifically, a grid search over models
with between 2 to 5 hidden layers, and
10 to 1000 hidden nodes. This MLP architecture
will be the same for all MLPs, and use ReLU activations. The latent dimensions $n_e$ and $n_h$ will also be searched in
the same space from 10 to 1000, independently of the hidden layer search.

We are interested if our allocation strategy with GNNs
produces better cosmological results than traditional methods.
Therefore, we will make a comparison with two alternatives: 1) We will employ the strategy commonly used in astronomy of uniform random selection among galaxies above a certain mass  or brightness threshold \citep{Levi2019-ws}; 2) We will adapt the evolutionary algorithm from \citep{Naghib2019-ko} to maximize \autoref{eq:utility} by varying the importance of a fixed set of well-motivated but predefined feature functions for policy proposals.
We define the metric for success as $\mathrm{var}^{-1}[\hat\varphi-\varphi_\mathrm{test}]$, where $\hat\varphi$ is computed for 50 test simulations with some $\varphi_\mathrm{test}\sim p(\varphi)$.

We are also interested in realistic applicability of this technique.
Therefore, we will measure the generalizability of this method to a slightly different simulator than the one used during training, and introduce biases in the data not seen in the training set, to determine if the learned observational strategy still produces better results than traditional methods.

\section{Documented Modifications}

We implemented the posterior noise model from \autoref{eq:posterior} as a MLP of the form $\phi^\Sigma(\mathbf{v}_i; r_i)\rightarrow (\Sigma_{d,i}, \Sigma_{\log m, i})$, i.e. a prediction model for the expected error after measurements with allocation $r_i$. Integration times are in the range 1-60 minutes. The advantage of this modification is twofold: 1) the function $\phi^\Sigma$ is routinely computed by astronomical surveys from simulated spectra; 2) the posterior model is differentiable unlike \autoref{eq:posterior}.

We changed the GNN architecture to include multiple messages as well as a global vector, using the full GN block as described in \cite{Battaglia2018-cy}. This change was required
for $\textit{GNN}_1$ to be capable of incorporating information about the entire field
before computing resource allocations. Otherwise, $\textit{GNN}_1$ would not be aware
of the total number of galaxies, and their properties, in the field. This
architecture was also used for $\textit{GNN}_2$.

We also chose to employ the publicly available genetic algorithm implementation \href{https://github.com/remiomosowon/pyeasyga}{\tt pyeasyga} instead of the one from \citep{Naghib2019-ko}.

\section{Results}
\label{sec:results}

\begin{figure}[t]
    \centering
    \subfigure{
        \includegraphics[width=0.5\textwidth]{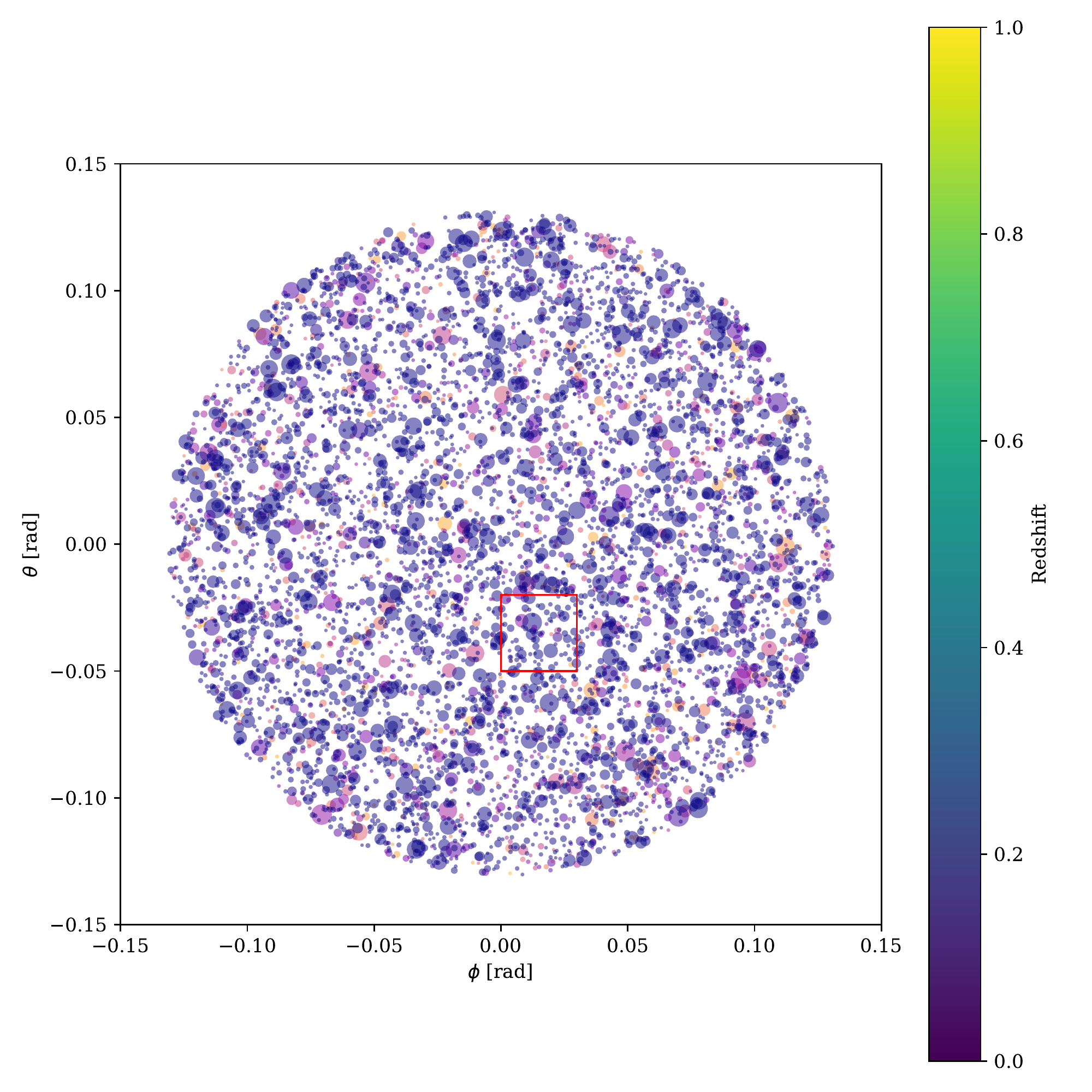}
    }%
    \subfigure{
        \includegraphics[width=0.5\textwidth]{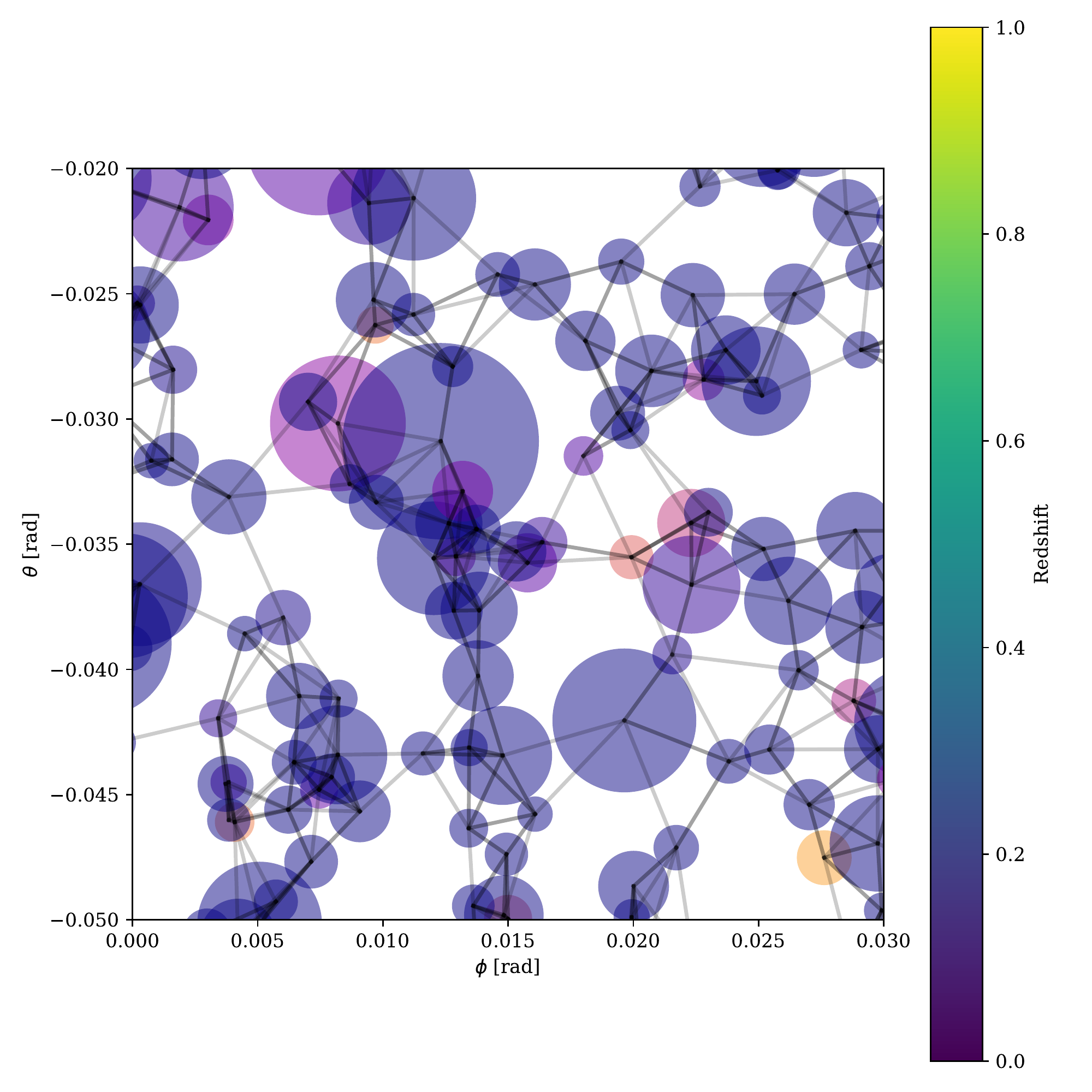}
    }
    \caption{A single field with approximately 2,000 galaxies obtained from a cosmological simulation (\emph{a}). Marker area indicates galaxy mass, marker color indicates distance from the observer. The graph (\emph{b}) for the boxed region in (\emph{a}).}
    \label{fig:fov}
\end{figure}

We use the cosmological simulations of \cite{2020ApJS..250....2V} for our experiments. 
There are 2,000 simulations for 2,000 different values of $\phi=\Omega_m$; with other cosmological parameters varied simultaneously by Latin hypercube sampling. For each simulation, we select a random direction on the sky and create a circular field with a radius of 7.5 degrees. 
For each field, we select uniformly at random a fraction of 0.001 of all galaxies and compute the graph for this galaxy list (see \autoref{fig:fov}); the subsampling permits faster training.
For each galaxy, we generate a simple parametric model for the spectrum, which incorporates the scaling of brightness with galaxy mass and distance.

\subsection{GNN Performance}

\begin{figure}[t]
    \centering
    \subfigure{
        \includegraphics[width=0.45\linewidth]{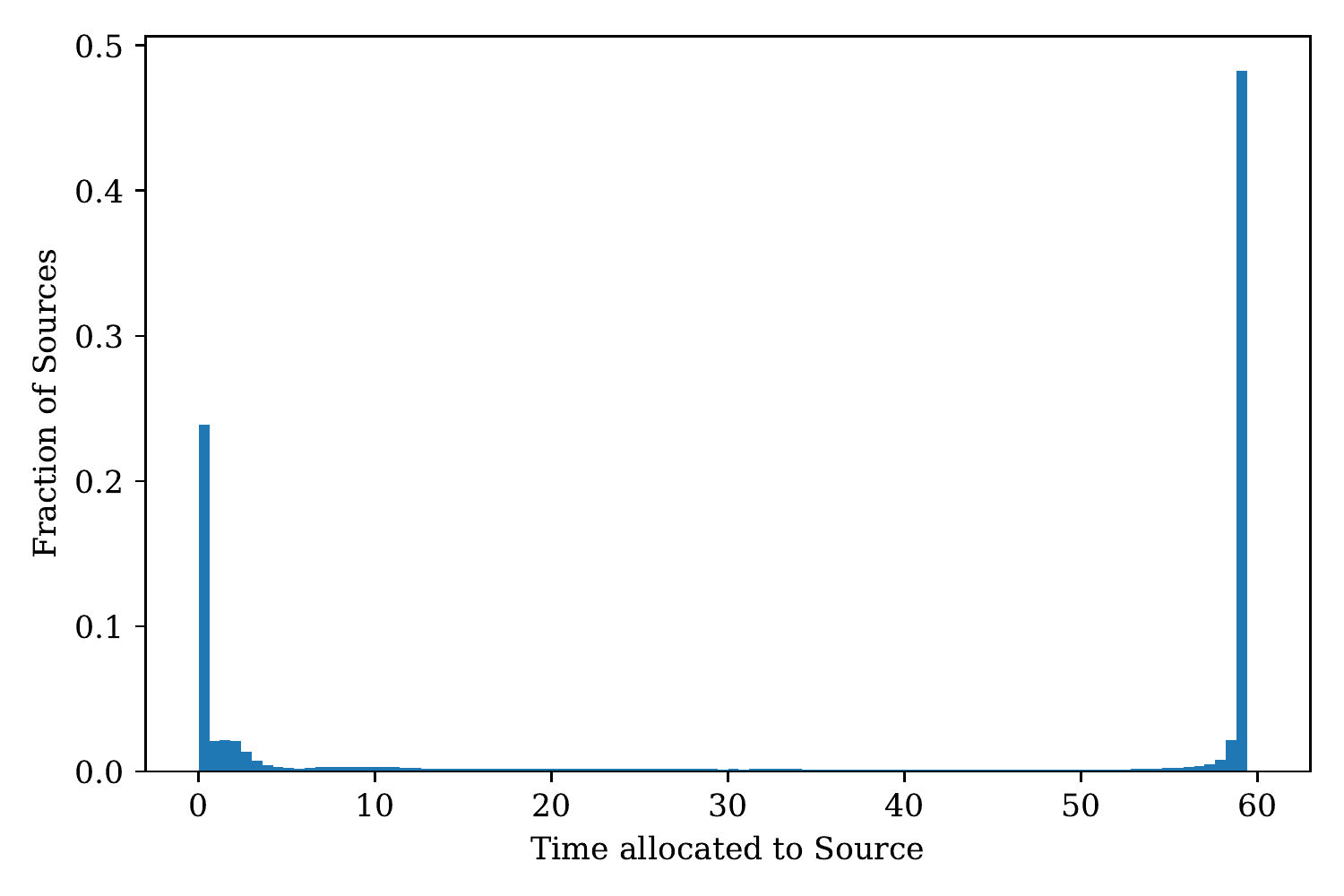}
    }
    \subfigure{
        \includegraphics[width=0.45\linewidth]{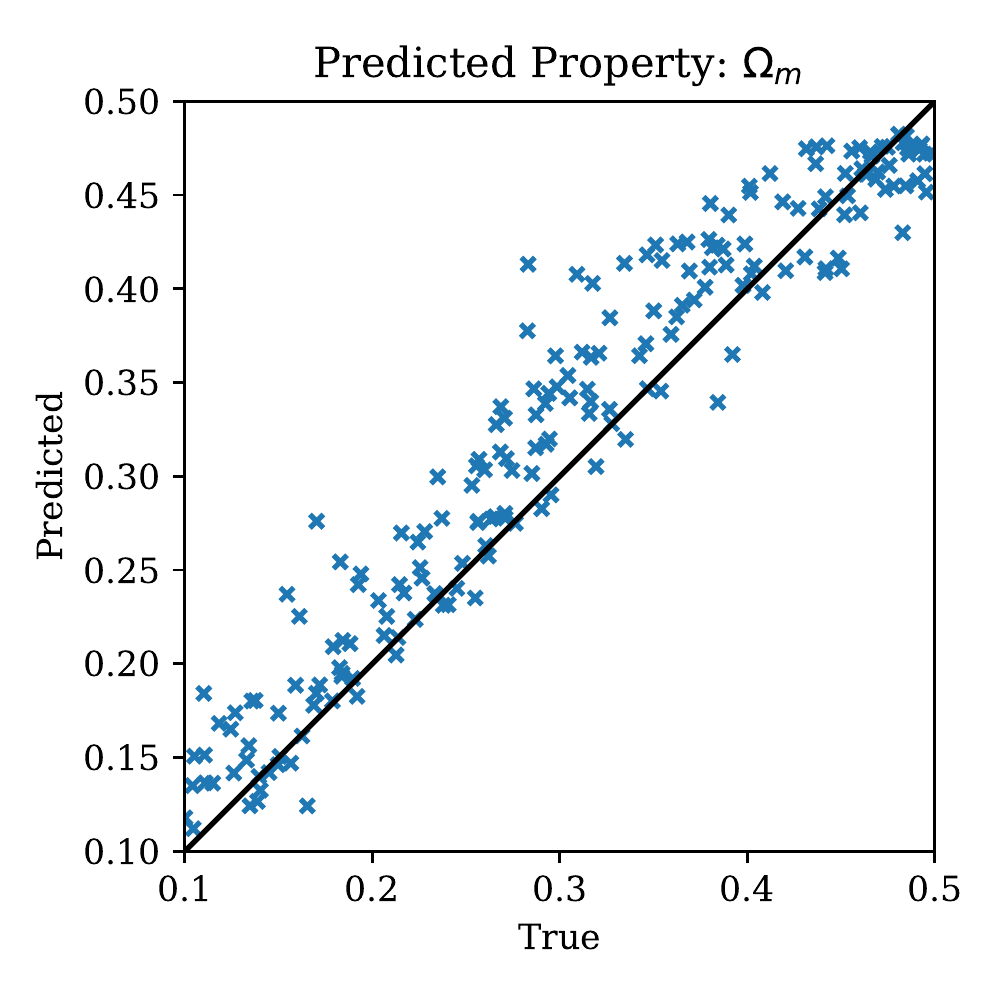}
    }
    \caption{Histogram of the allocated time per galaxy predicted by GNN$_1$ for a single test field (\emph{a}). Without any additional sparsity penalty, the network chose to allocate almost always either the minimum or maximum available time. $\Omega_m$ predictions from GNN$_2$ for a single field for each given test simulations (\emph{b}).}
    \label{fig:prediction}
\end{figure}

We use PyTorch and PyTorch Geometric to construct the GNNs and train them for 55 epochs on 1,800 distinct simulations.
The results of simultaneously optimizing the inference and allocation networks over these unknown simulated environments are shown in \autoref{fig:prediction}.
The GNN predicts $\Omega_m$ to better than 0.1 across the full range of the parameter, which is competitive with other methods of predicting this cosmological parameter from such a small set of galaxies.
As can be seen in the right panel of \autoref{fig:prediction}, there is a small positive bias for most parameter values, except at the high end, where there is a slight negative bias, which is a known consequence of a prior on $\Omega_m$ in the simulations we did not take into account.

\begin{figure}[t]
    \centering
    \includegraphics[width=\textwidth]{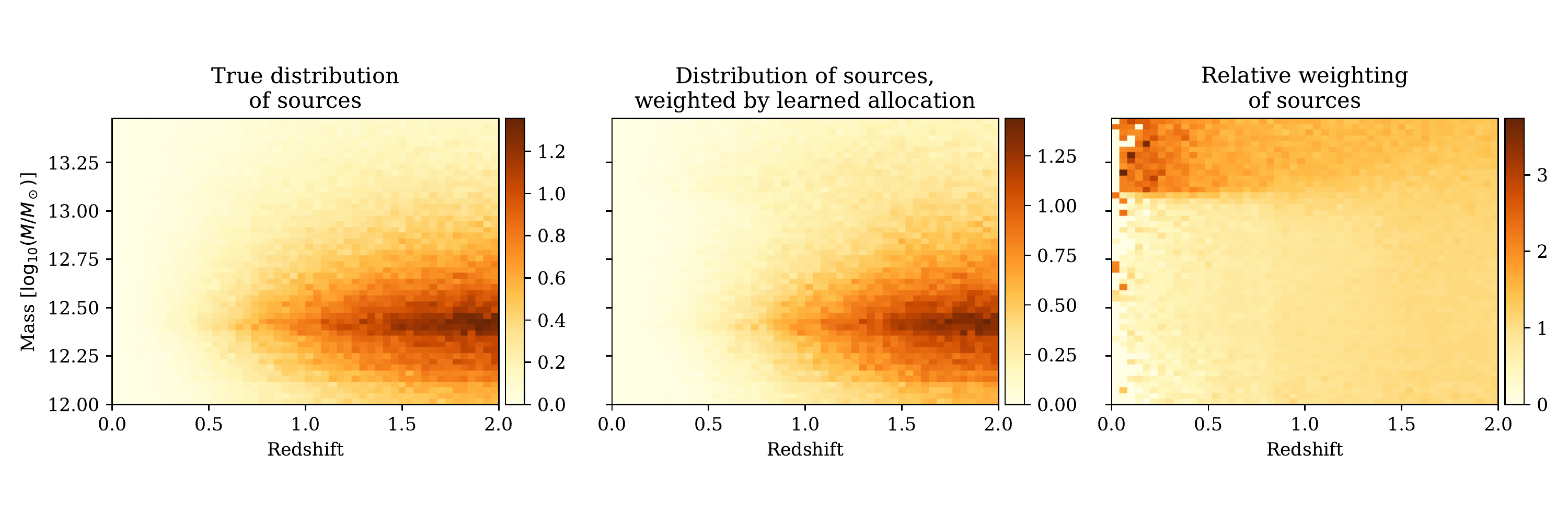}
    \caption{Learned allocations, visualized in mass-distance space.
    The left plot shows the true distribution of sources, given as a fractional number.
    The middle plot shows this distribution, weighted by resource allocation.
    The right plot shows the middle divided by the left, showing the allocation network
    favors nearby, high-mass galaxies.
    }
    \label{fig:sources}
\end{figure}

The time allocation budget is a compromise between the available resources and parameter prediction error.
Most galaxies in the simulated universes are far away and low in mass (see \autoref{fig:sources}, left panel).
Therefore, a long observation time is needed for each galaxy, as shown in the left panel of \autoref{fig:prediction}. 
However, the network can only select such galaxies until it runs out of time.
It therefore augments the result with a smaller number of low-mass, nearby galaxies, and selects them preferentially, presumably because they achieve good precision for the shortest possible integration time of 1 minute (\autoref{fig:sources}, right panel).

\subsection{Baseline Comparison with Genetic Algorithms}
\label{sec:baseline}

The baseline comparison implements a parameterized policy based on two relevant features in $\mathbf{v}$ --- the distance $d$ and the mass $\log m$.
For Baseline 1, we define, as is common in astronomical observations, a luminosity threshold $l_i > l_\mathrm{min}$, where $l_i\propto m_i / d_i^2$.
To decide how many resources are to be allocated to galaxy $i$, we implement the policy of maximizing the expected gain per unit resource, a known greedy policy for the unbounded knapsack problem \citep{Dantzig1957-ca}: 
\begin{equation}
\label{eq:marginal_gain}
    r_i = \mathrm{argmax}_r \Sigma_{d,i}^{-2}(r) / r.
\end{equation}
It is applied independently to each galaxy until the total allocation reaches the limit $H$.

For Baseline 2, we follow the idea of \citep{Naghib2019-ko} of using customized templates for the policy.
Specifically, we parameterize a candidate distance distribution as $d_\mathrm{cand} \sim \mathrm{Beta}(\alpha, \beta)$, and a mass distribution as $\log m_\mathrm{cand} \sim \mathrm{Beta}(\gamma + \delta d_\mathrm{cand}, 
\gamma + \delta (1-d_\mathrm{cand}))$.
The coupling between the two distributions encodes the physical necessity that selecting galaxies at larger distances $d$, at fixed observation time, requires them having higher masses.
We then match the candidate galaxies to the actually available galaxies by a nearest neighbor search in the space of $(d, \log m)$, and determine the proposed resources for each galaxy according to \autoref{eq:marginal_gain} until the allocation limit is reached.

For both policies, we optimize the parameters --- the luminosity threshold $l_\mathrm{min}$ or the distributional parameters $(\alpha,\beta,\gamma,\delta)$ --- by a genetic algorithm with 20 individuals, a mutation rate of 0.1, and at least 20 generations. 
To make the policies directly comparable to GNN$_1$, we set the fitness function to the prediction precision $\mathrm{var}^{-1}[\varphi-\hat\varphi]$ of GNN$_2$.
The results are summarized in \autoref{tab:performance}.

As we can see, the one-parameter Baseline 1 policy performs well, but not quite as well as our GNN$_1$.
This confirms the common practice in astronomy of selecting galaxies above a luminosity threshold as a simple and effective strategy.
Baseline 2 performs worse than Baseline 1. 
While the assumed Beta distributions are well intended and nominally capable of capturing the main trends of the mass and distance distributions of galaxies, they cannot follow the strongly non-uniform selection our GNN adopts and thus remain inferior.

\begin{table*}[]
    \centering
    \begin{tabular}{@{}l c c@{}}
    \toprule
    Method & Precision & Standard Deviation\\
    \midrule
    Allocation GNN & \bf{1117.39} & \bf{0.030}\\
    Baseline 1: Luminosity threshold &  828.46 & 0.035\\
    Baseline 2: Beta distribution & 496.23 & 0.045\\
    \bottomrule
    \end{tabular}
    \caption{Precision, i.e. inverse variance, and standard deviation of the inferred parameter $\Omega_m$ for our allocation network GNN$_1$ and two baseline models for allocation policies detailed in \autoref{sec:baseline}. The analysis network GNN$_2$ was used in all three cases to make predictions of $\Omega_m$ for 50 fields of a simulation with a constant value of $\Omega_m$.}
    \label{tab:performance}
\end{table*}

\section{Findings}
\label{sec:findings}

The presented implementation of an unsupervised GNN allocation algorithm achieves our intended goal of predicting a global environment parameter, $\Omega_m$, by simultaneously learning how to optimally allocate limited resources and how to analyze the results of that allocation.
It outperforms more conventional allocation strategies.
Our training occurs in an unknown environment with variations of the parameter(s) describing it.
This is distinct from and much more generally useful than traditional methods employed in astronomy, where a fixed and known environment is routinely assumed for forecasting purposes.
Furthermore, as evidenced by \autoref{fig:sources}, the allocation amounts to a decidedly non-uniform selection of galaxies, which traditional analysis methods needs to explicitly correct for, while our GNN$_2$ is automatically trained to do so. 
It appears likely that this capability of the analysis GNN to infer the underlying parameters despite a non-uniform selection naturally extends to many other analyses in astronomy that are strongly affected by non-uniform selections.
The presented study suggests that other applications in the observational sciences should similarly benefit from our flexible, unsupervised architecture of combining two GNNs to determine a strong allocation policy.

\small
\bibliography{references}

\begin{thebibliography}{26}
\providecommand{\natexlab}[1]{#1}
\providecommand{\url}[1]{\texttt{#1}}
\expandafter\ifx\csname urlstyle\endcsname\relax
  \providecommand{\doi}[1]{doi: #1}\else
  \providecommand{\doi}{doi: \begingroup \urlstyle{rm}\Url}\fi

\bibitem[Battaglia et~al.(2016)Battaglia, Pascanu, Lai, Rezende,
  et~al.]{battaglia2016interaction}
Peter Battaglia, Razvan Pascanu, Matthew Lai, Danilo~Jimenez Rezende, et~al.
\newblock {Interaction networks for learning about objects, relations and
  physics}.
\newblock In \emph{Advances in neural information processing systems}, pages
  4502--4510, 2016.

\bibitem[Battaglia et~al.(2018)Battaglia, Hamrick, Bapst, Sanchez-Gonzalez,
  Zambaldi, Malinowski, Tacchetti, Raposo, Santoro, Faulkner, Gulcehre, Song,
  Ballard, Gilmer, Dahl, Vaswani, Allen, Nash, Langston, Dyer, Heess, Wierstra,
  Kohli, Botvinick, Vinyals, Li, and Pascanu]{Battaglia2018-cy}
Peter~W Battaglia, Jessica~B Hamrick, Victor Bapst, Alvaro Sanchez-Gonzalez,
  Vinicius Zambaldi, Mateusz Malinowski, Andrea Tacchetti, David Raposo, Adam
  Santoro, Ryan Faulkner, Caglar Gulcehre, Francis Song, Andrew Ballard, Justin
  Gilmer, George Dahl, Ashish Vaswani, Kelsey Allen, Charles Nash, Victoria
  Langston, Chris Dyer, Nicolas Heess, Daan Wierstra, Pushmeet Kohli, Matt
  Botvinick, Oriol Vinyals, Yujia Li, and Razvan Pascanu.
\newblock Relational inductive biases, deep learning, and graph networks.
\newblock June 2018.
\newblock URL \url{http://arxiv.org/abs/1806.01261}.

\bibitem[Berthet et~al.(2020)Berthet, Blondel, Teboul, Cuturi, Vert, and
  Bach]{Berthet2020-co}
Quentin Berthet, Mathieu Blondel, Olivier Teboul, Marco Cuturi, Jean-Philippe
  Vert, and Francis Bach.
\newblock Learning with differentiable perturbed optimizers.
\newblock February 2020.
\newblock URL \url{http://arxiv.org/abs/2002.08676}.

\bibitem[Bronstein et~al.(2017)Bronstein, Bruna, LeCun, Szlam, and
  Vandergheynst]{bronstein2017geometric}
Michael~M Bronstein, Joan Bruna, Yann LeCun, Arthur Szlam, and Pierre
  Vandergheynst.
\newblock {Geometric deep learning: going beyond euclidean data}.
\newblock \emph{IEEE Signal Processing Magazine}, 34\penalty0 (4):\penalty0
  18--42, 2017.

\bibitem[Coles(2001)]{Coles2001-ql}
Peter Coles.
\newblock \emph{Cosmology: A Very Short Introduction}.
\newblock Oxford University Press, illustrated edition edition, December 2001.
\newblock ISBN 9780192854162.
\newblock URL
  \url{https://www.amazon.com/Cosmology-Short-Introduction-Peter-Coles/dp/019285416X}.

\bibitem[Cranmer et~al.(2020)Cranmer, Sanchez-Gonzalez, Battaglia, Xu, Cranmer,
  Spergel, and Ho]{cranmergn2}
Miles Cranmer, Alvaro Sanchez-Gonzalez, Peter Battaglia, Rui Xu, Kyle Cranmer,
  David Spergel, and Shirley Ho.
\newblock Discovering symbolic models from deep learning with inductive biases.
\newblock \emph{NeurIPS 2020}, 2020.

\bibitem[Cranmer et~al.(2019)Cranmer, Xu, Battaglia, and Ho]{cranmergn1}
Miles~D. Cranmer, Rui Xu, Peter Battaglia, and Shirley Ho.
\newblock Learning symbolic physics with graph networks.
\newblock \emph{Machine Learning and the Physical Sciences Workshop, NeurIPS
  2019}, 2019.

\bibitem[Dantzig(1957)]{Dantzig1957-ca}
George~B Dantzig.
\newblock {Discrete-Variable} extremum problems.
\newblock \emph{Operations research}, 5\penalty0 (2):\penalty0 266--288, April
  1957.
\newblock ISSN 0030-364X.
\newblock \doi{10.1287/opre.5.2.266}.
\newblock URL \url{https://doi.org/10.1287/opre.5.2.266}.

\bibitem[Eisen and Ribeiro(2020)]{gnnalloc2}
Mark Eisen and Alejandro Ribeiro.
\newblock Optimal wireless resource allocation with random edge graph neural
  networks.
\newblock \emph{IEEE Transactions on Signal Processing}, 68:\penalty0
  2977–2991, 2020.
\newblock ISSN 1941-0476.
\newblock \doi{10.1109/tsp.2020.2988255}.
\newblock URL \url{http://dx.doi.org/10.1109/TSP.2020.2988255}.

\bibitem[Gai and Qiu(2018)]{RL1}
Keke Gai and Meikang Qiu.
\newblock Optimal resource allocation using reinforcement learning for iot
  content-centric services.
\newblock \emph{Applied Soft Computing}, 70:\penalty0 12 -- 21, 2018.
\newblock ISSN 1568-4946.
\newblock \doi{https://doi.org/10.1016/j.asoc.2018.03.056}.
\newblock URL
  \url{http://www.sciencedirect.com/science/article/pii/S1568494618302540}.

\bibitem[Galstyan et~al.(2004)Galstyan, Czajkowski, and Lerman]{RL2}
Aram Galstyan, Karl Czajkowski, and Kristina Lerman.
\newblock Resource allocation in the grid using reinforcement learning.
\newblock In \emph{Proceedings of the Third International Joint Conference on
  Autonomous Agents and Multiagent Systems-Volume 3}, pages 1314--1315, 2004.

\bibitem[Gao et~al.(2020)Gao, Eisen, and Ribeiro]{gnnalloc}
Zhan Gao, Mark Eisen, and Alejandro Ribeiro.
\newblock Resource allocation via graph neural networks in free space optical
  fronthaul networks, 2020.

\bibitem[Gilmer et~al.(2017)Gilmer, Schoenholz, Riley, Vinyals, and
  Dahl]{gilmer2017neural}
Justin Gilmer, Samuel~S Schoenholz, Patrick~F Riley, Oriol Vinyals, and
  George~E Dahl.
\newblock Neural message passing for quantum chemistry.
\newblock In \emph{Proceedings of the 34th International Conference on Machine
  Learning-Volume 70}, pages 1263--1272. JMLR. org, 2017.

\bibitem[Lai et~al.(2020)Lai, Zha, Zhou, and Hu]{policygnn}
Kwei-Herng Lai, Daochen Zha, Kaixiong Zhou, and Xia Hu.
\newblock Policy-gnn: Aggregation optimization for graph neural networks.
\newblock In \emph{Proceedings of the 26th ACM SIGKDD International Conference
  on Knowledge Discovery \& Data Mining}, pages 461--471, 2020.

\bibitem[Levi et~al.(2019)Levi, Allen, Raichoor, Baltay, BenZvi, Beutler,
  Bolton, Castander, Chuang, Cooper, Cuby, Dey, Eisenstein, Fan, Flaugher,
  Frenk, Gonzalez-Morales, Graur, Guy, Habib, Honscheid, Juneau, Kneib, Lahav,
  Lang, Leauthaud, Lusso, de~la Macorra, Manera, Martini, Mao, Newman,
  Palanque-Delabrouille, Percival, Prieto, Rockosi, Ruhlmann-Kleider, Schlegel,
  Seo, Song, Tarle, Wechsler, Weinberg, Yeche, and Zu]{Levi2019-ws}
Michael~E Levi, Lori~E Allen, Anand Raichoor, Charles Baltay, Segev BenZvi,
  Florian Beutler, Adam Bolton, Francisco~J Castander, Chia-Hsun Chuang, Andrew
  Cooper, Jean-Gabriel Cuby, Arjun Dey, Daniel Eisenstein, Xiaohui Fan, Brenna
  Flaugher, Carlos Frenk, Alma~X Gonzalez-Morales, Or~Graur, Julien Guy, Salman
  Habib, Klaus Honscheid, Stephanie Juneau, Jean-Paul Kneib, Ofer Lahav, Dustin
  Lang, Alexie Leauthaud, Betta Lusso, Axel de~la Macorra, Marc Manera, Paul
  Martini, Shude Mao, Jeffrey~A Newman, Nathalie Palanque-Delabrouille, Will~J
  Percival, Carlos~Allende Prieto, Constance~M Rockosi, Vanina
  Ruhlmann-Kleider, David Schlegel, Hee-Jong Seo, Yong-Seon Song, Greg Tarle,
  Risa Wechsler, David Weinberg, Christophe Yeche, and Ying Zu.
\newblock The dark energy spectroscopic instrument ({DESI}).
\newblock July 2019.
\newblock URL \url{http://arxiv.org/abs/1907.10688}.

\bibitem[Liu et~al.(2017)Liu, Li, Xu, Xu, Lin, Qiu, Tang, and Wang]{RL3}
Ning Liu, Zhe Li, Zhiyuan Xu, Jielong Xu, Sheng Lin, Qinru Qiu, Jian Tang, and
  Yanzhi Wang.
\newblock A hierarchical framework of cloud resource allocation and power
  management using deep reinforcement learning, 2017.

\bibitem[Naghib et~al.(2019)Naghib, Yoachim, Vanderbei, Connolly, and
  Lynne~Jones]{Naghib2019-ko}
Elahesadat Naghib, Peter Yoachim, Robert~J Vanderbei, Andrew~J Connolly, and
  R~Lynne~Jones.
\newblock A framework for telescope schedulers: With applications to the large
  synoptic survey telescope.
\newblock \emph{The Astronomical Journal}, 157\penalty0 (4):\penalty0 151,
  March 2019.
\newblock ISSN 0002-9602, 1538-3881.
\newblock \doi{10.3847/1538-3881/aafece}.
\newblock URL \url{http://dx.doi.org/10.3847/1538-3881/aafece}.

\bibitem[Scarselli et~al.(2009)Scarselli, Gori, Tsoi, Hagenbuchner, and
  Monfardini]{scarselli2009graph}
Franco Scarselli, Marco Gori, Ah~Chung Tsoi, Markus Hagenbuchner, and Gabriele
  Monfardini.
\newblock The graph neural network model.
\newblock \emph{IEEE Transactions on Neural Networks}, 20\penalty0
  (1):\penalty0 61--80, 2009.

\bibitem[Silver et~al.(2016)Silver, Huang, Maddison, Guez, Sifre, van~den
  Driessche, Schrittwieser, Antonoglou, Panneershelvam, Lanctot, Dieleman,
  Grewe, Nham, Kalchbrenner, Sutskever, Lillicrap, Leach, Kavukcuoglu, Graepel,
  and Hassabis]{Silver2016-qe}
David Silver, Aja Huang, Chris~J Maddison, Arthur Guez, Laurent Sifre, George
  van~den Driessche, Julian Schrittwieser, Ioannis Antonoglou, Veda
  Panneershelvam, Marc Lanctot, Sander Dieleman, Dominik Grewe, John Nham, Nal
  Kalchbrenner, Ilya Sutskever, Timothy Lillicrap, Madeleine Leach, Koray
  Kavukcuoglu, Thore Graepel, and Demis Hassabis.
\newblock Mastering the game of go with deep neural networks and tree search.
\newblock \emph{Nature}, 529\penalty0 (7587):\penalty0 484--489, January 2016.
\newblock ISSN 0028-0836, 1476-4687.
\newblock \doi{10.1038/nature16961}.
\newblock URL \url{http://dx.doi.org/10.1038/nature16961}.

\bibitem[Storn and Price(1997)]{Storn1997-he}
Rainer Storn and Kenneth Price.
\newblock Differential evolution -- a simple and efficient heuristic for global
  optimization over continuous spaces.
\newblock \emph{Journal of Global Optimization}, 11\penalty0 (4):\penalty0
  341--359, December 1997.
\newblock ISSN 0925-5001, 1573-2916.
\newblock \doi{10.1023/A:1008202821328}.
\newblock URL \url{https://doi.org/10.1023/A:1008202821328}.

\bibitem[Tabibian et~al.(2020)Tabibian, G{\'o}mez, De, Sch{\"o}lkopf, and
  Gomez~Rodriguez]{Tabibian2020-cn}
Behzad Tabibian, Vicen{\c c} G{\'o}mez, Abir De, Bernhard Sch{\"o}lkopf, and
  Manuel Gomez~Rodriguez.
\newblock On the design of consequential ranking algorithms.
\newblock volume 124 of \emph{Proceedings of Machine Learning Research}, pages
  171--180, Virtual, 2020. PMLR.
\newblock URL \url{http://proceedings.mlr.press/v124/tabibian20a.html}.

\bibitem[Villaescusa-Navarro et~al.(2020)Villaescusa-Navarro, Hahn, Massara,
  Banerjee, Delgado, Ramanah, Charnock, Giusarma, Li, Allys, Brochard,
  Uhlemann, Chiang, He, Pisani, Obuljen, Feng, Castorina, Contardo, Kreisch,
  Nicola, Alsing, Scoccimarro, Verde, Viel, Ho, Mallat, Wandelt, and
  Spergel]{Villaescusa-Navarro2020-wa}
Francisco Villaescusa-Navarro, Changhoon Hahn, Elena Massara, Arka Banerjee,
  Ana~Maria Delgado, Doogesh~Kodi Ramanah, Tom Charnock, Elena Giusarma, Yin
  Li, Erwan Allys, Antoine Brochard, Cora Uhlemann, Chi-Ting Chiang, Siyu He,
  Alice Pisani, Andrej Obuljen, Yu~Feng, Emanuele Castorina, Gabriella
  Contardo, Christina~D Kreisch, Andrina Nicola, Justin Alsing, Roman
  Scoccimarro, Licia Verde, Matteo Viel, Shirley Ho, Stephane Mallat, Benjamin
  Wandelt, and David~N Spergel.
\newblock The quijote simulations.
\newblock \emph{The Astrophysical Journal Supplement Series}, 250\penalty0
  (1):\penalty0 2, August 2020.
\newblock ISSN 0067-0049.
\newblock \doi{10.3847/1538-4365/ab9d82}.
\newblock URL
  \url{https://iopscience.iop.org/article/10.3847/1538-4365/ab9d82/meta}.

\bibitem[{Villaescusa-Navarro} et~al.(2020){Villaescusa-Navarro}, {Hahn},
  {Massara}, {Banerjee}, {Delgado}, {Ramanah}, {Charnock}, {Giusarma}, {Li},
  {Allys}, {Brochard}, {Uhlemann}, {Chiang}, {He}, {Pisani}, {Obuljen}, {Feng},
  {Castorina}, {Contardo}, {Kreisch}, {Nicola}, {Alsing}, {Scoccimarro},
  {Verde}, {Viel}, {Ho}, {Mallat}, {Wandelt}, and
  {Spergel}]{2020ApJS..250....2V}
Francisco {Villaescusa-Navarro}, ChangHoon {Hahn}, Elena {Massara}, Arka
  {Banerjee}, Ana~Maria {Delgado}, Doogesh~Kodi {Ramanah}, Tom {Charnock},
  Elena {Giusarma}, Yin {Li}, Erwan {Allys}, Antoine {Brochard}, Cora
  {Uhlemann}, Chi-Ting {Chiang}, Siyu {He}, Alice {Pisani}, Andrej {Obuljen},
  Yu~{Feng}, Emanuele {Castorina}, Gabriella {Contardo}, Christina~D.
  {Kreisch}, Andrina {Nicola}, Justin {Alsing}, Roman {Scoccimarro}, Licia
  {Verde}, Matteo {Viel}, Shirley {Ho}, Stephane {Mallat}, Benjamin {Wandelt},
  and David~N. {Spergel}.
\newblock {The Quijote Simulations}.
\newblock \emph{\apjs}, 250\penalty0 (1):\penalty0 2, September 2020.
\newblock \doi{10.3847/1538-4365/ab9d82}.

\bibitem[Vlastelica et~al.(2019)Vlastelica, Paulus, Musil, Martius, and
  Rol{\'\i}nek]{Vlastelica2019-np}
Marin Vlastelica, Anselm Paulus, V{\'\i}t Musil, Georg Martius, and Michal
  Rol{\'\i}nek.
\newblock Differentiation of blackbox combinatorial solvers.
\newblock December 2019.
\newblock URL \url{http://arxiv.org/abs/1912.02175}.

\bibitem[Wang et~al.(2018)Wang, Liao, Ba, and Fidler]{nervenet}
Tingwu Wang, Renjie Liao, Jimmy Ba, and Sanja Fidler.
\newblock Nervenet: Learning structured policy with graph neural networks.
\newblock In \emph{International Conference on Learning Representations}, 2018.

\bibitem[{Ye} et~al.(2019){Ye}, {Li}, and {Juang}]{8633948}
H.~{Ye}, G.~Y. {Li}, and B.~F. {Juang}.
\newblock Deep reinforcement learning based resource allocation for v2v
  communications.
\newblock \emph{IEEE Transactions on Vehicular Technology}, 68\penalty0
  (4):\penalty0 3163--3173, 2019.

\end{thebibliography}

\end{document}